\documentclass[letterpaper, 10 pt, conference]{ieeeconf}  
\usepackage{graphicx}
\usepackage{subfig}
\usepackage{dirtree}
\usepackage{multirow}
\usepackage{hyperref}

\IEEEoverridecommandlockouts                              

\title{\LARGE \bf
How to Count Coughs: An Event-Based Framework for Evaluating Automatic Cough Detection Algorithm Performance
}

\author{Lara Orlandic,$^{1}$ Jonathan Dan,$^{1}$ Jérôme Thevenot,$^{1}$ Tomas Teijeiro$^{2}$, Alain Sauty$^{3}$, and David Atienza$^{1}$ 
\thanks{$^{1}$Embedded Systems Laboratory (ESL) of EPFL, Lausanne, Switzerland. Corresponding author: {\tt\small lara.orlandic@epfl.ch}}
\thanks{$^{2}$BCAM - Basque Center for Applied Mathematics, Bilbao, Spain}
\thanks{$^{3}$Lausanne University Hospital (CHUV), Lausanne, Switzerland}%
}

\begin{document}

\maketitle
\thispagestyle{empty}
\pagestyle{empty}

\begin{abstract}

Chronic cough disorders are widespread and challenging to assess because they rely on subjective patient questionnaires about cough frequency. Wearable devices running Machine Learning (ML) algorithms are promising for quantifying daily coughs, providing clinicians with objective metrics to track symptoms and evaluate treatments. However, there is a mismatch between state-of-the-art metrics for cough counting algorithms and the information relevant to clinicians. Most works focus on distinguishing cough from non-cough samples, which does not directly provide clinically relevant outcomes such as the number of cough events or their temporal patterns. In addition, typical metrics such as specificity and accuracy can be biased by class imbalance. We propose using event-based evaluation metrics aligned with clinical guidelines on significant cough counting endpoints. We use an ML classifier to illustrate the shortcomings of traditional sample-based accuracy measurements, highlighting their variance due to dataset class imbalance and sample window length. We also present an open-source event-based evaluation framework to test algorithm performance in identifying cough events and rejecting false positives. We provide examples and best practice guidelines in event-based cough counting as a necessary first step to assess algorithm performance with clinical relevance. 

\end{abstract}

\section{INTRODUCTION}

Chronic cough and cough hypersensitivity disorders are globally prevalent conditions that significantly impair patients' quality of life. These conditions are difficult to treat due to the difficulty in identifying causes, including individual triggers and underlying pulmonary disorders~\cite{won_impact_2021}. Current clinical practice assesses severity and treatment efficacy through patient questionnaires, which are only moderately correlated to actual cough counts~\cite{turner_measuring_2023}. Hence, there is significant interest in using smart wearable devices to automatically provide objective daily cough counts as a more accurate, unbiased means of assessment~\cite{turner_measuring_2023, birring_leicester_2008, hall_present_2020, serrurier_past_2022}.

The guidelines of the European Respiratory Society (ERS)~\cite{morice_ers_2007} highlight multiple clinically significant endpoints in cough monitoring, including: 1) the number of cough events, 2) seconds containing at least one cough, 3) breaths followed by a cough, and 4) cough bouts. Studying the pattern of coughing is crucial, as cough bouts correlate more closely than individual coughs with pathology and reported severity, and can indicate different underlying physiological mechanisms~\cite{dockry_s17_2022}. Thus, automated tools should monitor both cough frequency and the temporal distribution of cough patterns to provide insights into the patient's symptomatology and guide treatment plans~\cite{turner_measuring_2023}.

The rise in multi-parametric wearable devices has enabled the development of automatic cough counting tools using sensors and ML classifiers~\cite{hall_present_2020, serrurier_past_2022}. However, there is a gap between reported algorithm performance metrics and clinically relevant endpoints. Metrics such as specificity (SP) are often reported, but may misrepresent practicality as they are heavily influenced by cough frequency and long periods of silence, and therefore do not contribute useful information in a long-term monitoring scenario ~\cite{drugman_objective_2013}. Furthermore, most studies use fixed-length windows, typically on the order of seconds, for features extraction and classification~\cite{drugman_objective_2013, otoshi_novel_2021}. This time granularity cannot accurately count individual cough events or distinguish cough patterns, as a cough typically lasts 0.3-0.5~s~\cite{serrurier_past_2022}.

This work analyzes the interpretation of algorithm performance in the context of cough counting, implementing common evaluation metrics from the literature to assess their robustness to methodological choices such as dataset imbalance and inference window length. Using an audio-based reference classifier, we demonstrate how the performance of a single algorithm can vary widely depending on the chosen metrics and experimental setup. Finally, we propose an event-based (EB) validation framework for fair and meaningful evaluation of cough counting algorithms, focusing on detecting individual cough events to provide detailed information about cough bouts.

\section{METHODS}

\subsection{Methodological choices in experimental setup}

Algorithm performance can be influenced by a number of methodological choices in experiment design, such as the noise and class imbalance present in the data, as well as the signal window length used in training and inference. Instead of using one fixed set of choices to evaluate different ML model architectures, we showed that the same algorithm trained with the same data can yield widely varying results depending on the choice of setup and performance metrics.

\subsubsection{Dataset class imbalance}

In this work, we use the only known open dataset to delineate the start and stop times of each individual cough sound \cite{orlandic_multimodal_2023}. It contains audio and kinematic data from healthy subjects performing forced coughs, breathing, throat clearing, speaking, and laughing. The data is collected under scenarios of audio and kinematic noise, such as traffic noise or subject movement, to assess the robustness of the algorithm to such conditions.

\begin{figure}[ht]
  \centering
  \includegraphics[width=0.8\linewidth]{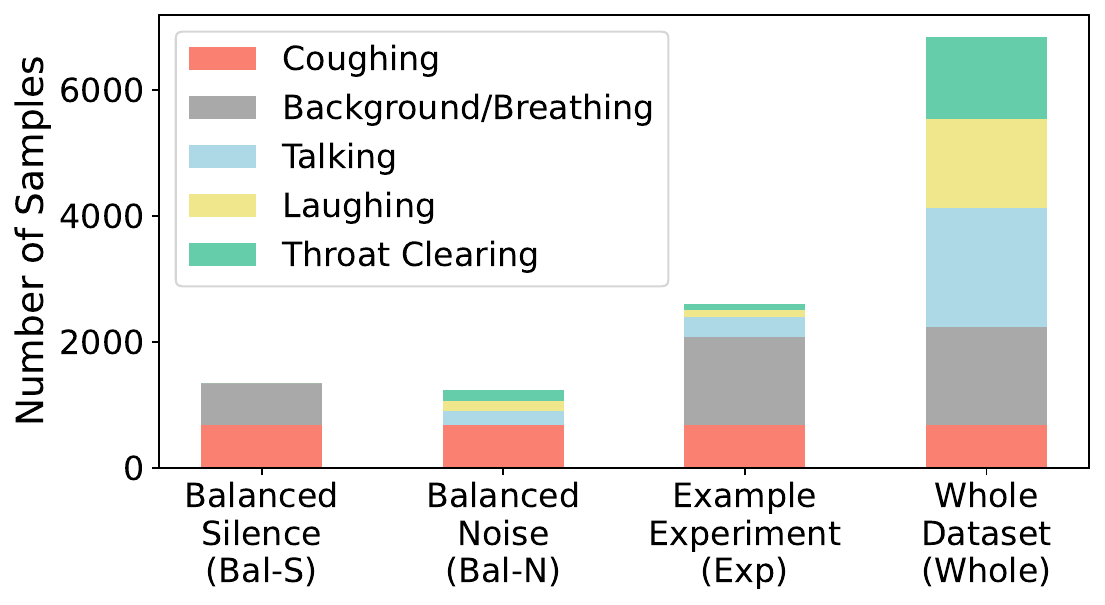} 
\caption{Test dataset distributions with varying class imbalance}
\label{fig:test_sets}
\end{figure}

We use signals from a chest-worn microphone to train and test the audio-based classifier described in Section~\ref{sec:ml}. To illustrate the behavior of various metrics in different class imbalance scenarios, we sample the test dataset to match four distributions with different proportions of cough, silence, and non-cough noise recordings. Each scenario contains the same number of cough recordings, but the non-cough ones are randomly eliminated to match the target distribution. These recordings are then segmented into 0.8 s windows to obtain the distributions of samples shown in Fig.~\ref{fig:test_sets}. The first scenario, (Bal-S), is a balanced dataset of coughing and background noise or breathing. In the balanced noise (Bal-N) scenario, the cough samples are balanced with sounds that can be confused for coughing. The third scenario, example experiment (Exp) approximates the data distributions reported by a State-of-the-Art cough counting study \cite{drugman_objective_2013}. Finally, the Whole Dataset (Whole) scenario uses all the available segmented testing data, in which the distributions of each sound are roughly equal.

\subsubsection{Window length}

To extract physiological features on a wearable device, sensor data is processed in windows of a given length. A ML classifier is then run with a given overlap between windows. During the data segmentation phase, window length is used to determine which samples are counted as cough versus non-cough samples, with True Positive (TP) cough samples defined as windows that contain a whole or fraction of a cough event. However, in sample-based (SB) models, longer windows can hinder the ability of a classifier to detect individual cough events. For example, Otoshi \textit{et al.} use 5 s windows, and therefore 10 sequential coughs over 10 s would be counted as 2 coughs \cite{otoshi_novel_2021}. 

We further investigate the influence of window length on the ability of classifiers to count coughs by training and testing classifiers using lengths ranging from 0.4 s to 1 s with 0.1 s increments. Each model is re-trained for each given window length, and then tested on the whole test dataset. We do not consider the effects of window overlap on the model performance in an online testing scenario, as it can lead to the same coughs being counted multiple times and consequently skewed TP values. We discuss how this issue is solved with EB signal post-processing in Section ~\ref{sec:ml}.

\subsection{Sample-based (SB) evaluation metrics}

The most prevalent algorithm performance metrics in the cough counting literature are Sensitivity (SE) and Specificity (SP), reported in 85\% and 62\% of works, respectively. Furthermore, less than 50\% of works report Accuracy (AC), Precision (PR), F-1 score (F1), ROC-AUC, Negative Predictive value (NPV), and False positives per hour (FP/hr). 
SB models count TPs, True Negatives (TNs), False Positives (FPs), and False Negatives (FNs) at the sample level depending on whether a given sample of data contains at least one cough or not. We report each of these SB metrics for different dataset distributions to illustrate their relevance, or lack thereof, in an online wearable cough-counting scenario.

\subsection{Event-based (EB) performance metrics}
\label{sec:event_based_metrics}

\begin{figure}[ht]
  \centering
  \includegraphics[width=\linewidth]{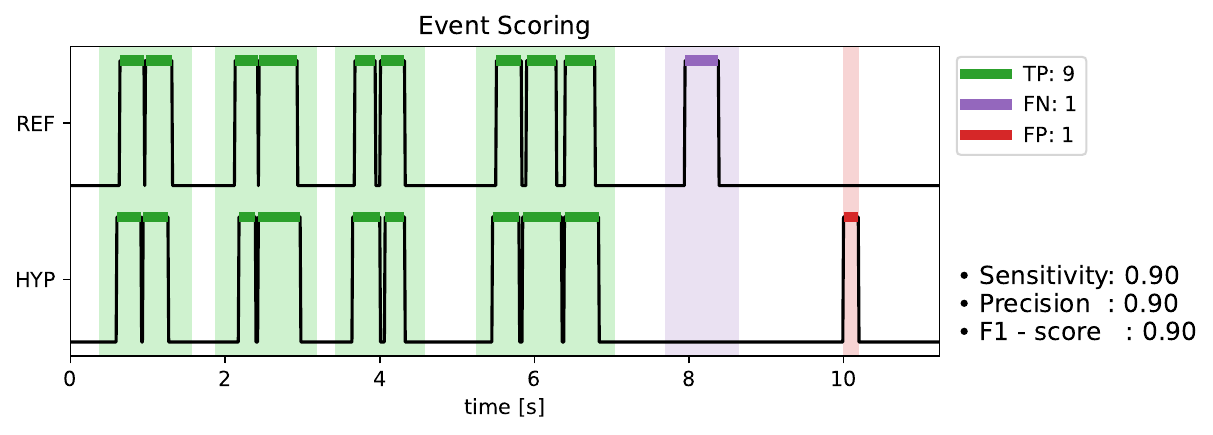} 
\caption{An example of the EB scoring library evaluating the performance of a classifier's prediction}
\label{fig:scoring}
\end{figure}

Unlike SB metrics, EB metrics measure the ability of an algorithm to correctly identify the temporal locations of annotated events. TPs, TNs, FPs, and FNs are determined based on the overlap between individual true and predicted cough events. Such EB metrics are a classic approach used in long-term monitoring systems, such as detecting epileptic seizures \cite{dan_szcore_2024}, or polyphonic sounds \cite{mesaros_metrics_2016}.

The EB performance metrics have been computed on each test dataset using the timescoring library \cite{dan_szcore_2024}, an example of which is shown in Fig. \ref{fig:scoring}. In order to adapt the code to our cough counting task, certain parameters of the scoring needed to be modified, namely the maximum event duration and tolerance thresholds. These parameters were set based on cough physiology \cite{chang_physiology_2006}; coughs usually last up to 0.5 s, so a maximum event duration of 0.6 s was chosen to account for vocal artefacts following a cough. Any longer event is automatically segmented into multiple events.

\begin{figure}[b]
  \centering
  \includegraphics[width=0.8\linewidth]{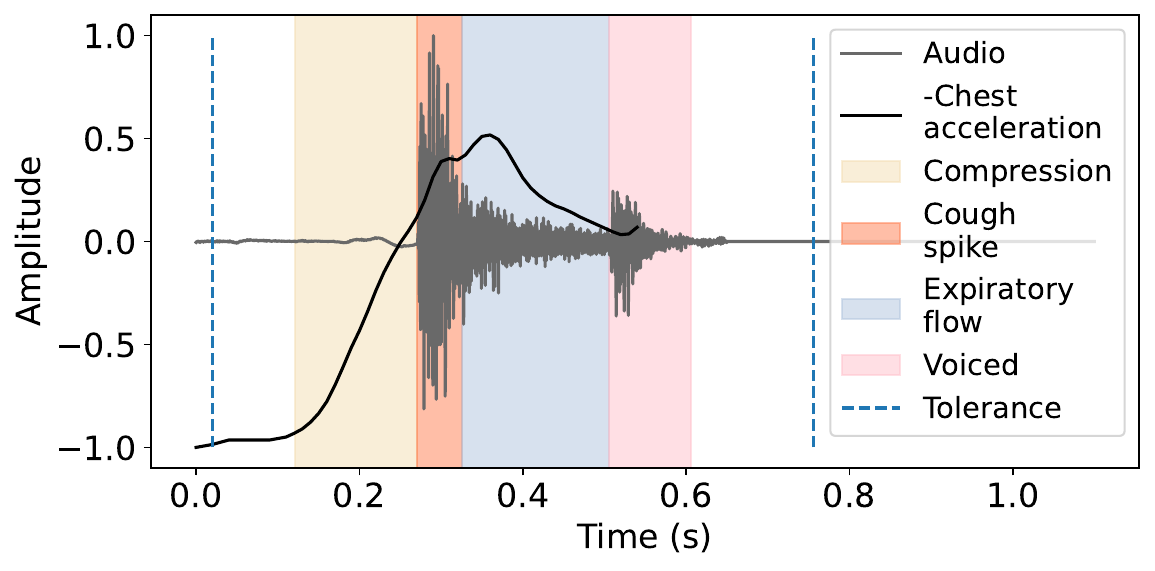} 
\caption{Delineated phases of a cough episode.}
\label{fig:cough_delineation}
\end{figure}

\begin{figure*}[ht]
  \centering
  \includegraphics[width=0.8\textwidth]{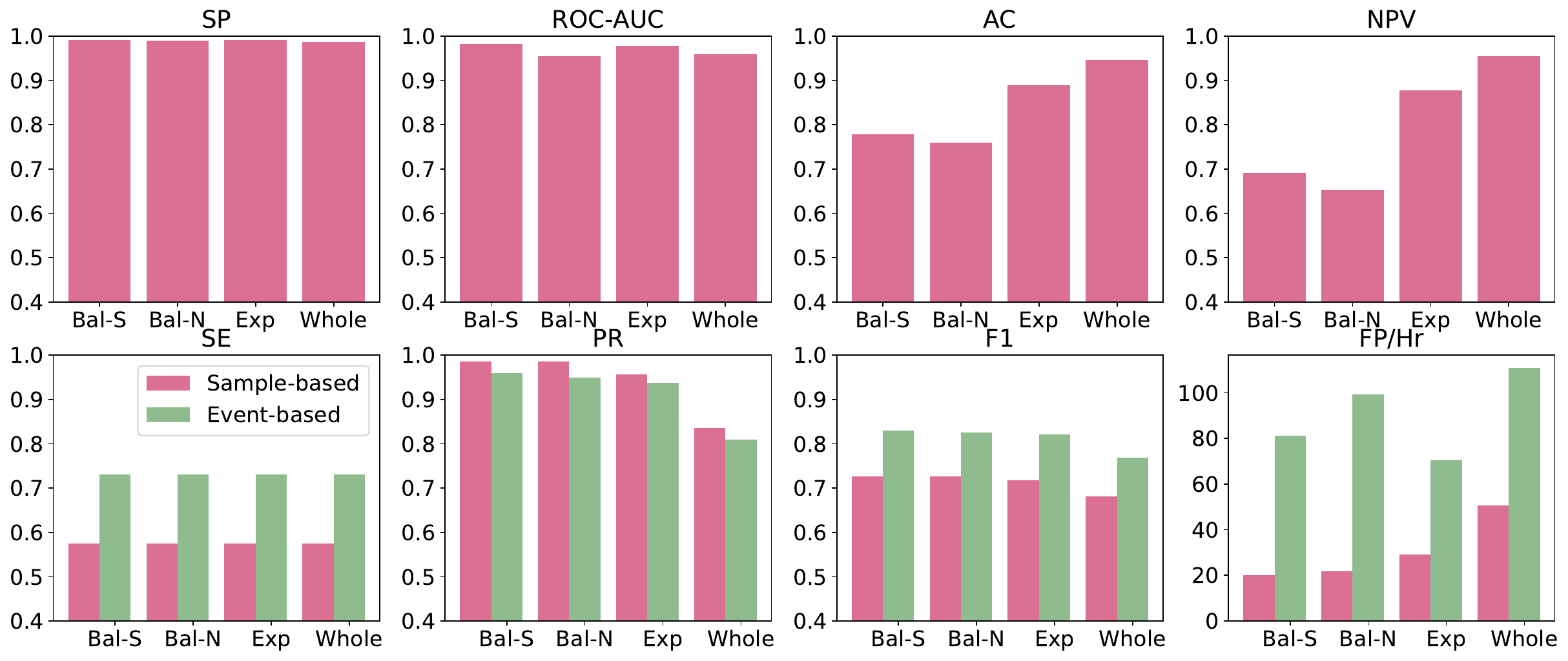} 
\caption{Performance metrics extracted on different test dataset distributions for the SB and EB ML models.}
\label{fig:acc_seg}
\vspace{-1em}
\end{figure*}

Fig.~\ref{fig:cough_delineation} shows an example delineation of the different phases of a cough sound. It starts with a compressive phase of approximately 0.2 s in which pressure builds in the lungs. Consequently, the chest compresses and movement can be observed in the acceleration that is normal to the chest. The compressive phase is followed by an explosive cough spike of 0.03-0.05 s that accounts for a loud noise in the cough audio signal. Next, air exits the lungs in the expiratory phase, lasting 0.2-0.5 s \cite{chang_physiology_2006}. Finally, some cough sounds contain a voiced component whose duration is unspecified. Consequently, start and end tolerance values, which determine how much deviation in the predicted and true event locations is tolerated when counting TPs, were set to 0.25 s. This value corresponds to the sum of the compressive and cough spike phases, as well as the cough spike and expiratory phases. These tolerance values are shown with respect to the ground-truth cough onset and offset. An example of computing EB performance metrics of a cough counting prediction has been added to our \href{https://github.com/esl-epfl/edge-ai-cough-count/}{public Git repository}.

\subsection{Machine Learning training and testing}
\label{sec:ml}
A simple ML classifier was developed to illustrate the differences in model performance across different scenarios and metrics. For each incoming data sample, we first compute the same audio features that have been previously used in cough classification \cite{orlandic_semi-supervised_2023}. To enhance the generalizability of the algorithm to unseen test subjects and balance the data classes, the training dataset is augmented in a semi-supervised fashion with cough samples from the COUGHVID dataset as performed in \cite{orlandic_semi-supervised_2023}. Finally, the augmented training dataset is used to train an eXtreme Gradient Boosting (XGB) classifier.

The same classifier is used to compute the SB and EB metrics. We have added a novel post-processing algorithm inspired by cough physiology to detect individual cough events. When the classifier detects a cough window, the algorithm identifies cough spike regions by applying hysteresis thresholding on the signal power. The classifier and thresholding procedure are run with a 50\% window overlap to ensure that all cough spikes are detected despite changes in the signal baseline. Once the regions are identified for multiple windows, the algorithm checks that the identified cough locations are physiologically possible by merging regions whose peaks are too close. Finally, the algorithm adapts the start and end locations of the coughs, based on the acceptable ranges of each phase shown in Fig. ~\ref{fig:cough_delineation}, as well as the subject's average cough duration from previous cough bursts. This produces a series of non-overlapping regions for each predicted cough event.

\section{RESULTS}

\subsection{Relevance of performance metrics}

The SB accuracy results of testing the same ML model on four different dataset distributions are shown in Fig.~\ref{fig:acc_seg}. The first conclusion drawn from this figure is that several metrics vary greatly depending on the testing dataset distribution, even though the different datasets are subsets of the same whole dataset. SE is constant because all testing dataset scenarios use the same number of cough samples. However, PR decreases by 15.1\% from Bal-S to Whole due to the increased presence of non-cough samples and consequent FPs. This is also shown by a 2.5x increase in FP/hr between Bal-S and the Whole datasets in the SB model.

Next, we observe shortcomings in the 4 metrics based on TN samples, namely SP, AC, ROC-AUC, and NPV. Fig.~\ref{fig:acc_seg} illustrates that AC and NPV increase as the percentage of cough samples decreases in the dataset. Thus, these scores are higher in the case of imbalanced data and do not reflect how well the classifier functions, as the most challenging dataset in terms of non-cough noises (i.e. Whole) exhibits the best performance by these metrics. Furthermore, SP and its derivative AUC are high in all scenarios because the number of TNs of the classifier vastly outweighs the number of FPs, thus saturating the metrics to high values regardless of the scenario. As the purpose of the wearable device is to accurately detect coughs with as few FPs and FNs as possible, the number of TNs is irrelevant to the performance of the device. Therefore, cough counting models should be evaluated in terms of SE, PR, F1, and FP/hr.

\subsection{SB vs. EB scoring}

The EB model described in Section~\ref{sec:ml} is tested on the same  test data distributions using the scoring methodology in Section~\ref{sec:event_based_metrics}, and the results are shown in Fig.~\ref{fig:acc_seg}. Similarly to the SB metrics, the Whole dataset generates the highest number of FPs and consequently the model's F1 is 6.5\% lower in the Whole scenario than that of Exp. 
This figure also illustrates the utility of the FP/hr metric. The PR of the Bal-N scenario is 1.2\% higher than that of the Exp one, meaning that the ratio of FPs to TPs is lower in Bal-N. However, we see that the FP/hr of Bal-N is 28.9\% higher than that of Exp. This is because there are more total recordings in the Exp scenario, many of which are silent, and therefore there are fewer FPs per recording time. Ideally, both metrics should be reported to fully explore the FP rate of the algorithm.

\begin{figure}[t]
  \centering
  \includegraphics[width=0.8\linewidth]{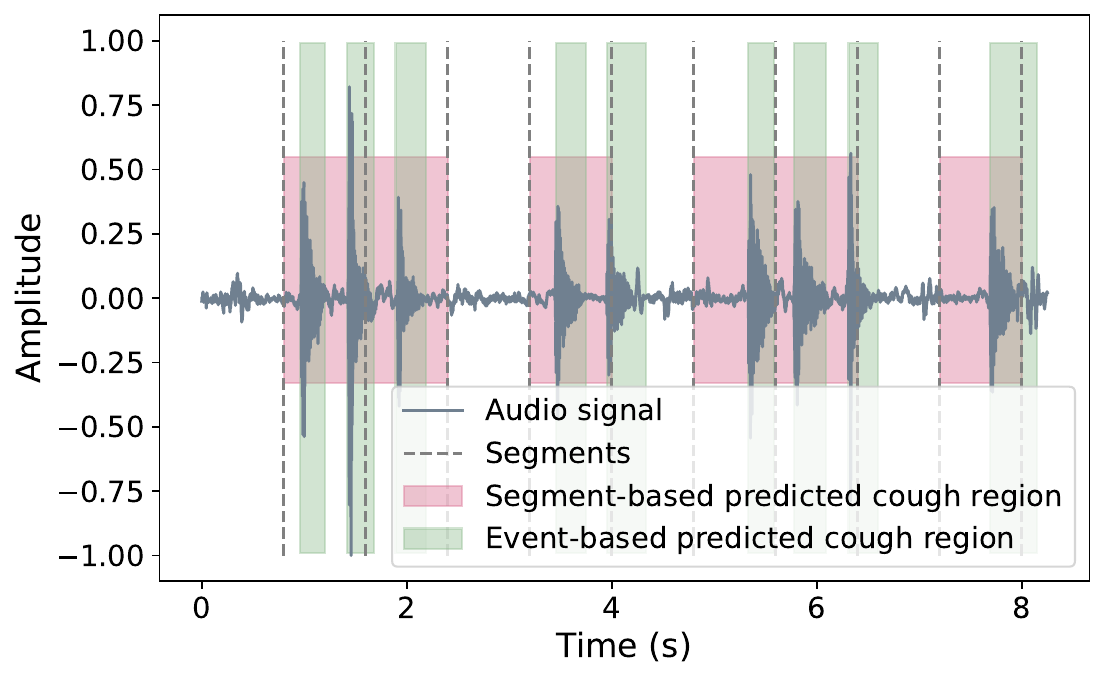} 
\caption{Example of the SB model and enhanced EB detector delineating a series of cough bursts.}
\label{fig:samp_vs_event}
\vspace{-1.5em}
\end{figure}

The SE of the EB model is 73.1\% as opposed to 57.4\% in the SB classifier, perhaps because the overlap of the EB model can better detect missed coughs, as for the fifth cough in Fig.~\ref{fig:samp_vs_event}. This figure shows the SB and EB classifier outputs on an example test recording. We can see that the SB model only identifies regions in which coughs are present, while the EB model provides information about the pattern in which the coughs occur, whether there are one or multiple coughs in a region, how long each cough lasts, and much more.

Finally, we investigate the effects of varying the window length of the classifier in the SB and EB models. Although the total of TPs does not vary much with window length in the EB model (st. dev: 23 coughs), the TPs steadily decrease with window length in the SB classifier (st. dev: 95 coughs). This is because the SB model can only guarantee one cough per window, and a longer window includes more coughs. Therefore, as observed by Otoshi \textit{et al.} \cite{otoshi_novel_2021}, the number of coughs gets further distorted as the window length increases. In contrast, the EB model's segmentation algorithm identifies each individual event within a window and breaks up events too long to be a single cough.

\section{DISCUSSION AND CONCLUSIONS}

Chronic cough requires personalized monitoring, and automated cough counting algorithms on wearables can provide objective information about the performance of individual treatment plans. However, to compare algorithmic approaches, the research community requires a clear definition of performance metrics that are relevant to this context. Traditional metrics like Specificity and Accuracy, used in over 60\% of studies, are highly sensitive to class imbalance present in the dataset. We strongly recommend using Sensitivity, Precision, F-1 Score, and False Positives per hour for evaluation.

Next, this work proposes a clinically relevant methodology for evaluating cough event detection in line with the ERS guidelines. Traditional segmentation and ML classifiers fail to accurately count coughs or distinguish between isolated and consecutive cough bouts. Thus, we propose an event-based evaluation framework to identify individual cough onsets and offsets, providing precise information on cough patterns to assess disease severity and treatment efficacy. We also demonstrate how to transform a sample-based classifier into an event-based model using a novel, physiology-inspired post-processing algorithm. We provide examples of event-based testing on \href{https://github.com/esl-epfl/edge-ai-cough-count/}{our public GitHub repository} for the research community to freely extend to their datasets and analyses. This strategy can be applied to monitor other respiratory disorder symptoms (i.e., wheezing, stridor).

Finally, our analysis shows that even event-based metrics can vary with the distribution of the test dataset. Details about the dataset, including its cough-to-non-cough ratio and non-cough sample contents, must be clearly stated in performance claims. A limitation of this work and much of the state-of-the-art is that the algorithms are trained on distributions of forced coughs by healthy subjects. An open, labeled, clinical dataset is needed to compare the performance of the algorithm in patients with chronic cough disorders.

\addtolength{\textheight}{0cm}   




\section*{ACKNOWLEDGMENT}

This work has been partially supported by the EU’s Horizon 2020 grant agreement no. 101017915 (DIGIPREDICT). T. Teijeiro is supported by the grant RYC2021-032853-I funded by MCIN/AEI and by NextGenerationEU/PRTR.


\bibliographystyle{IEEEtran}
\bibliography{IEEEabrv, refs}

\begin{thebibliography}{10}
\providecommand{\url}[1]{#1}
\csname url@samestyle\endcsname
\providecommand{\newblock}{\relax}
\providecommand{\bibinfo}[2]{#2}
\providecommand{\BIBentrySTDinterwordspacing}{\spaceskip=0pt\relax}
\providecommand{\BIBentryALTinterwordstretchfactor}{4}
\providecommand{\BIBentryALTinterwordspacing}{\spaceskip=\fontdimen2\font plus
\BIBentryALTinterwordstretchfactor\fontdimen3\font minus \fontdimen4\font\relax}
\providecommand{\BIBforeignlanguage}[2]{{%
\expandafter\ifx\csname l@#1\endcsname\relax
\typeout{** WARNING: IEEEtran.bst: No hyphenation pattern has been}%
\typeout{** loaded for the language `#1'. Using the pattern for}%
\typeout{** the default language instead.}%
\else
\language=\csname l@#1\endcsname
\fi
#2}}
\providecommand{\BIBdecl}{\relax}
\BIBdecl

\bibitem{won_impact_2021}
H.-K. Won \emph{et~al.}, ``Impact and disease burden of chronic cough,'' \emph{Asia Pacific Allergy}, vol.~11, no.~2, p. e22, 2021.

\bibitem{turner_measuring_2023}
R.~D. Turner \emph{et~al.}, ``Measuring cough: what really matters?'' \emph{Journal of Thoracic Disease}, vol.~15, no.~4, pp. 2288--2299, Apr. 2023.

\bibitem{birring_leicester_2008}
S.~S. Birring \emph{et~al.}, ``The {Leicester} {Cough} {Monitor}: preliminary validation of an automated cough detection system in chronic cough,'' \emph{The European Respiratory Journal}, vol.~31, no.~5, 2008.

\bibitem{hall_present_2020}
J.~I. Hall \emph{et~al.}, ``The present and future of cough counting tools,'' \emph{Journal of Thoracic Disease}, vol.~12, no.~9, pp. 5207--5223, 2020.

\bibitem{serrurier_past_2022}
A.~Serrurier \emph{et~al.}, ``\BIBforeignlanguage{en}{Past and {Trends} in {Cough} {Sound} {Acquisition}, {Automatic} {Detection} and {Automatic} {Classification}: {A} {Comparative} {Review}},'' \emph{\BIBforeignlanguage{en}{Sensors}}, vol.~22, no.~8, p. 2896, Jan. 2022.

\bibitem{morice_ers_2007}
A.~H. Morice \emph{et~al.}, ``\BIBforeignlanguage{en}{{ERS} guidelines on the assessment of cough},'' \emph{\BIBforeignlanguage{en}{European Respiratory Journal}}, vol.~29, no.~6, 2007.

\bibitem{dockry_s17_2022}
R.~J. Dockry \emph{et~al.}, ``\BIBforeignlanguage{en}{S17 {A} relevant definition of cough bouts},'' \emph{\BIBforeignlanguage{en}{Thorax}}, vol.~77, no. Suppl 1, pp. A14--A15, Nov. 2022.

\bibitem{drugman_objective_2013}
T.~Drugman \emph{et~al.}, ``Objective {Study} of {Sensor} {Relevance} for {Automatic} {Cough} {Detection},'' \emph{IEEE Journal of Biomedical and Health Informatics}, vol.~17, no.~3, pp. 699--707, May 2013.

\bibitem{otoshi_novel_2021}
T.~Otoshi \emph{et~al.}, ``\BIBforeignlanguage{en}{A novel automatic cough frequency monitoring system combining a triaxial accelerometer and a stretchable strain sensor},'' \emph{\BIBforeignlanguage{en}{Scientific Reports}}, vol.~11, 2021.

\bibitem{orlandic_multimodal_2023}
L.~Orlandic \emph{et~al.}, ``\BIBforeignlanguage{en}{A {Multimodal} {Dataset} for {Automatic} {Edge}-{AI} {Cough} {Detection}},'' in \emph{\BIBforeignlanguage{en}{IEEE EMBC}}, Sydney, Jul. 2023, pp. 1--7.

\bibitem{dan_szcore_2024}
J.~Dan \emph{et~al.}, ``{SzCORE}: {A} {Seizure} {Community} {Open}-source {Research} {Evaluation} framework for the validation of {EEG}-based automated seizure detection algorithms,'' Mar. 2024.

\bibitem{mesaros_metrics_2016}
A.~Mesaros \emph{et~al.}, ``\BIBforeignlanguage{en}{Metrics for {Polyphonic} {Sound} {Event} {Detection}},'' \emph{\BIBforeignlanguage{en}{Applied Sciences}}, vol.~6, no.~6, p. 162, Jun. 2016.

\bibitem{chang_physiology_2006}
A.~B. Chang, ``\BIBforeignlanguage{en}{The physiology of cough},'' \emph{\BIBforeignlanguage{en}{Paediatric Respiratory Reviews}}, vol.~7, no.~1, pp. 2--8, Mar. 2006.

\bibitem{orlandic_semi-supervised_2023}
L.~Orlandic \emph{et~al.}, ``A semi-supervised algorithm for improving the consistency of crowdsourced datasets: {The} {COVID}-19 case study on respiratory disorder classification,'' \emph{Computer Methods and Programs in Biomedicine}, vol. 241, p. 107743, Nov. 2023.

\end{thebibliography}

\end{document}